\documentclass[applsci,article,accept,moreauthors,pdftex]{Definitions/mdpi} 

\firstpage{1} 
\pubvolume{1}
\issuenum{1}
\articlenumber{0}
\pubyear{2021}
\copyrightyear{2021}
\externaleditor{Academic Editor: Antonio Fernández-Caballero} 
\datereceived{23 March 2021} 
\dateaccepted{7 April 2021} 
\datepublished{} 
\hreflink{https://doi.org/} 

\Title{Bio-Inspired Modality Fusion for Active Speaker Detection}

\TitleCitation{Bio-Inspired Modality Fusion for Active Speaker Detection}

\Author{Gustavo Assunção $^{1,2,}$*$^{,\dagger}$\orcidA{}, Nuno Gonçalves $^{1,2,\dagger}$\orcidB{} and Paulo Menezes $^{1,2,\dagger}$\orcidC{}}

\AuthorNames{Gustavo Assunção, Nuno Gonçalves and Paulo Menezes}

\AuthorCitation{Assuncao, G.; Goncalves, N.; Menezes, P.}

\address{%
$^{1}$ \quad University of Coimbra, Department of Electrical and Computer Engineering; nunogon@isr.uc.pt (N.G.); pm@deec.uc.pt (P.M.)\\
$^{2}$ \quad Institute of Systems and Robotics, 3030-194 Coimbra, Portugal}

\corres{Correspondence: gustavo.assuncao@isr.uc.pt}

\firstnote{Current address: ISR, R. Silvio Lima, 3030-194 Coimbra, Portugal} 

\abstract{Human beings have developed fantastic abilities to integrate information from various sensory sources exploring their inherent complementarity. Perceptual capabilities are therefore heightened, enabling, for instance, the well-known "cocktail party" and McGurk effects, i.e., speech disambiguation from a panoply of sound signals. This fusion ability is also key in refining the perception of sound source location, as in distinguishing whose voice is being heard in a group conversation. Furthermore, neuroscience has successfully identified the superior colliculus region in the brain as the one responsible for this modality fusion, with a handful of biological models having been proposed to approach its underlying neurophysiological process. Deriving inspiration from one of these models, this paper presents a methodology for effectively fusing correlated auditory and visual information for active speaker detection. Such an ability can have a wide range of applications, from teleconferencing systems to social robotics. The detection approach initially routes auditory and visual information through two specialized neural network structures. The resulting embeddings are fused via a novel layer based on the superior colliculus, whose topological structure emulates spatial neuron cross-mapping of unimodal perceptual fields. The validation process employed two publicly available datasets, with achieved results confirming and greatly surpassing initial expectations.}


\begin{document}

\section{Introduction}

    The ability to discern between activity and passivity in humans may well benefit social robots and autonomous cognition systems by enabling a more accurate and natural response in interactive scenarios. Accordingly, machines capable of this task would be able to measure a user's engagement level and infer valuable conclusions. Moreover, successfully solving this task would expedite a research shift towards other problems whose solutions depend on the output of a human activity--passivity recognizer. Recent work on data mining and collection \cite{nagrani} and speaker diarization \cite{Ephrat}, which heavily rely on the use of active speaker detection, are exemplary of the topic's importance.
    
\begin{figure}[h]
        \begin{center}
            \includegraphics[width=\columnwidth]{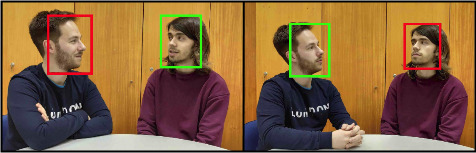}
        \end{center}
            \caption{Exemplary output of an multi-speaker active speaker detection (ASD) system where an intervening speaker is denoted in green and a passive speaker is encircled in red.}
            \label{fig:example}
    \end{figure}

    Active speaker detection (ASD), or the ability to recognize who in a group is speaking at any given time, is a subset of human activity--passivity recognition. This subset focuses on mapping utterances that make up a speech signal, such as in a conversation, to either a closed or open set of intervening speakers. In turn, techniques from the subset may take advantage of audio-only, visual-only or audiovisual information depending on their nature. Though studies do exist taking advantage of other types of perceptual input, such as \cite{Vajaria}, these are far more uncommon and somewhat less successful. In the past, extensive research effort has been put towards the audio-only section of ASD as a side tool for other problems \cite{Tranter,Bouquin}. Techniques generally relying on sequential speaker clustering \cite{Liu} or speech source localization \cite{Maraboina,Bertrand},  were often enough to achieve real-time diarization, some occasionally aided by supervised learning. Unfortunately these were constrained by predetermined assumptions---mic location, static speakers, controlled environment. Furthermore, audio-only approaches remain unable to match segmented and clustered speech with visual speakers without additional data or restrictions, limiting their use on audiovisual data. Hence, visual approaches to the topic become useful, though not without their own drawbacks. Unreservedly interpreting lip movement \cite{Siatras}, or other facial feature variations \cite{Ahmad,Stefanov},  as vocal activity without accounting for similar motions stemming from normal human interaction---emotional expression, coughing, chewing---will of course negatively impact the detection process. An exemplary output of an active speaker detection system is shown in Figure~\ref{fig:example}, for easier understanding.

    In terms of multi-modal approaches, by means of complementing a visual approach with its audio counterpart or vice-versa, performances are generally higher thanks to increased robustness. This improvement is obtained by exploiting correlations between the two types of data (e.g., \cite{Minotto,Cutler}, which may decrease ambiguous situations prone to cause model confusion---speaker overlap, unrelated movement. Self-evidently, techniques requiring both a visual and audio input are unable to deal with situations where either of the modalities is missing. These situations however are scarce, an do not correspond to the immensely vast amount of audiovisual data available nowadays. Thus multi-modality remains as the preferred manner of addressing active speaker detection \cite{Chakravarty,Chakravarty_2,Stefanov_2}.
    
    In spite of the recent technological advancements which have allowed deep learning research to flourish, such improvements would likely not have been possible without key biological contributions to the area. The success of remarkable architectures such as convolutional neural networks, whose earliest iterations \cite{Fukushima} strived to emulate neuron behavior in the visual cortex of primates \cite{Hubel}, would likely have been halted without this biological inspiration. However, engineering maintains a collaborative yet stubbornly segregated relationship with biology by disregarding several other promising models which the latter has to offer.
  
    In this study our contribution is two-fold as we propose a new multi-modal approach to visual ASD as well as a novel bio-inspired modality fusion layer which strives to emulate the superior colliculus structure of the brain. The approach is designed to work in real-time and makes no prior assumptions regarding noise, number of participants or spatial configuration. This is because it was developed also with the intent of being used in parallel projects, where it is required to recognize speakers and the moments they intervene in. The newly designed modality fusion layer performs successful integration of multi-source uni-sensory information through feedback stimulation of spatially proximal neural regions. The obtained results were highly positive, having exceeded expectations in terms of current state-of-the-art standards.
    
    The presented paper is structured as follows. Initially, examination and insight into previous research is shown in Section 2, introducing the reader into the topic and contextualizing our approach. The proposed technique is described in Section 3, followed by an overview of the experiments carried out in Section 4 and the obtained results. These are then examined and critically discussed in Section 5. Finally, a conclusion is provided in Section 6.
 
\section{Related Work}

    The overview presented is branched into two sub-sections. To begin with, recent research on the topic is summarized in order to give further insight into ASD, and contextualize our own approach to the matter. Following that, we analyze a number of the few available datasets and their suitability as an integrating part of ASD techniques.
    
    \subsection{ASD Recent Research}
    Given the broad array of early audio-centered approaches, and their lingering success with nonvisual data, current efforts have turned more towards vision-oriented issues. This is not to say audio has stopped being a key aspect of current approaches to ASD. In fact, multi-modal state-of-the-art techniques have used machine learning to realize audiovisual correlations in data. Such techniques are successful, for instance, as a means to recognize speech related facial motion while also discerning it from other unrelated movements. \mbox{In \cite{Stefanov_2,Stefanov_3}},  Stefanov {et al.} used the outputs of a face detector, along with audio-derived voice activity detection (VAD) labels, to train a CNN task specific feature extractor together with a Perceptron classifier. For transfer learning comparison, known CNN architectures were used as pre-trained feature extractors whose outputs were employed in training temporal (LSTM net) and nontemporal (Perceptron) classifiers. In a related work \cite{Ren}, Ren {et al.} tackled clutter and unrelated motion fallible situations by extending conventional LSTM models to share learned weights across the audio and visual modalities. Thus learning audiovisual correlations to improve model robustness. With a real implementation in a NAO robot, Cech {et al.} \cite{Cech} used time difference of arrival (TDOA) over a set of microphone pairs to simultaneously evince potential sound sources in 2D space. A facial detector then estimated mouth locations in 3D space, to be statistically mapped to the former 2D sound sources and TDOAs. In a similar fashion, Gebru {et al.} \cite{Gebru} mapped located sound sources to the 2D image plane as well as estimated potential speaker lip positions. However, this cross-modal weighting approach employed a GMM variation to assign weights to audio observations based on their spatial proximity to visual observations, and vice-versa. Audiovisual clustering then achieved active speaker detection. To give another example, Hoover {et al.} \cite{Hoover} attempted correlating the degree of occurrence of individual faces in a video with single speaker clusters of speech segments, obtained using vectors of locally aggregated descriptors (VLADs) \cite{Jegou}. 
    
    In addition to the usual reliance on high-end equipment or ideal settings (i.e. consistent number of speakers, frontal facing and nonoverlapping speech, mic/camera location) which is common to clustering and other methods, further issues also occur with current state-of-the-art techniques. Particularly, unidirectional dependency between modalities is recurrent among machine learning approaches. The lack of carefully labeled big data on both the audio and visual modalities for ASD evidently forces resourcing to alternatives which transfer labels generated from one modality to its sibling (typically from audio to video). Such processes unavoidably lead to propagation of error, originating from the single modality prediction techniques, which could be avoided were large ASD datasets existent and applicable to robust machine learning architectures.

    \subsection{Datasets}
    In order to validate audiovisual ASD techniques (and train when required), some data encompassing all considered modalities must be available. Furthermore, such data is required to include some form of labeling so as to enable its application in the aforementioned techniques. Unfortunately, the multi-modal nature and challenges inherent to audiovisual ASD constitute a major setback in terms of acquiring labeled data. As such, small and constrained datasets are frequently built by research teams just to evaluate their own approaches, which of course restricts their comparability and reduces the credibility of their success. Yet very recently a few datasets have been created and made available, which address this current necessity in ASD research.
    
    In \cite{Ephrat} the {AVSpeech} dataset was presented, an automatically collected large-scale corpus made up of several lecture recordings available on YouTube. These recordings, amounting to around 4700 h of audiovisual data, always show a single clearly visible face to which the ongoing speech corresponds. This dataset could be labeled for active speaker detection using some voice activity detection (VAD) or speech-silence discerning technique, and then applied for training and testing of ASD approaches. However, this may lead to model confusion later on in multi-speaker scenarios and as such {AVSpeech} is perhaps best left for ASD testing. A potentially better example of a suitable corpus is given in \cite{Chakravarty_2}, where the {Columbia} dataset was first introduced. Here, the recording of a panel discussion between seven speakers was overlapped with upper body bounding boxes for all participants, each labeled with speak/no-speak markers. By always including more than one speaker in each frame, this dataset has the benefit over {AVSpeech} to enable dealing with multi-speaker situations in inference models. As for its downside, the dataset is of reduced size and its data is considerably homogeneous. This evidently means it may be harder to extrapolate relevant features only and generalize a model trained on this data, which constrains the dataset's usage in machine learning approaches.

    The largest and most heterogeneous active speaker detection dataset available at the moment is presumably {AVA-ActiveSpeaker} \cite{Roth}, developed by Google for the 4th ActivityNet challenge at CVPR 2019. This hand labeled set of segments obtained from 160 YouTube videos amounts to around 38.5 h of audiovisual data, where each of the covered \mbox{3.65 million} frames is detailed with bounding box locations for all detected speaker faces as well as with three types of labels for each present bounding box---{speaking and audible, speaking but not audible, not speaking}. Such a corpus should therefore be perfect for learning ASD as it covers all previously mentioned requirements for a suitable audiovisual dataset and presents no major drawbacks. To support this, the team behind {AVA-ActiveSpeaker} used it to train and test both static and recurrent prediction models, applying modality fusion and consistently achieving great performances. Unfortunately however, this dataset encompasses an extremely high amount of low quality videos in terms of image resolution, audio intelligibility and noise. The creators of the dataset even went so far as to include several voice-over videos and dubbed versions of movies not in their original language. This naturally means a copious amount of audio clips are out of sync with their supposedly respective facial image sequences. Additionally, the lengths of these clips frequently do not pair with the lengths of those facial sequences---due to phrases varying in duration for different languages. Taking this into account each of the 160 videos was manually inspected for quality assessment, as though they may be appropriate for a challenge in the same conditions, it would be senseless to use such great amounts of noisy incorrect data to train a model and expect it to perform successfully with correct data and in the real world. A total of 20 videos were considered to meet today's audiovisual capture standards. 

\section{Methodology}

    This section describes the path taken to develop the proposed technique aimed at performing active speaker detection, including detailed descriptions of each individual component of the overall system.

    The developed method is capable of dealing with either audio data, visual data or audiovisual data. This is because there is an independent feature extraction module for each modality, with resulting embeddings being assigned significance by the fusion module only if they are provided. Additionally, it can also deal with multiple speaker scenarios as it processes each person individually. In order to achieve this, audiovisual raw data is divided into audio segments and its corresponding sequences of facial images. Narrowband spectrograms are generated for the audio segments and propagated through the {VGGVox} model \cite{nagrani}, while features and corresponding temporal correlations are obtained from the facial sequences using a modified {ResNet50} architecture \cite{He} with added recurrence (henceforth called RN-LSTM). The convolutional stage of the former has been pretrained with the {VGGFace2} dataset \cite{Cao} and is available in \cite{Keras_vggface}. Detail is further provided for each model component in the following sections.

    \subsection{Bio-Inspired Embedding Fusion}
    
        Integration of multi-modal information is a considerably complex task characterized by deeply convoluted data correlations. In the animal kingdom, such a task is unconsciously performed on certain brain regions, allowing for beings and equally humans to derive multi-sensory conclusions. Inspired by the work of Ursino {et al.} in \cite{Ursino} and their model of the brain's superior colliculus (SC) for multi-sensory integration, we present a novel bio-inspired neural network layer for audio and visual embedding fusion. This section describes its structure and behavior.
        
\begin{figure}[H]
            \includegraphics[width=\columnwidth]{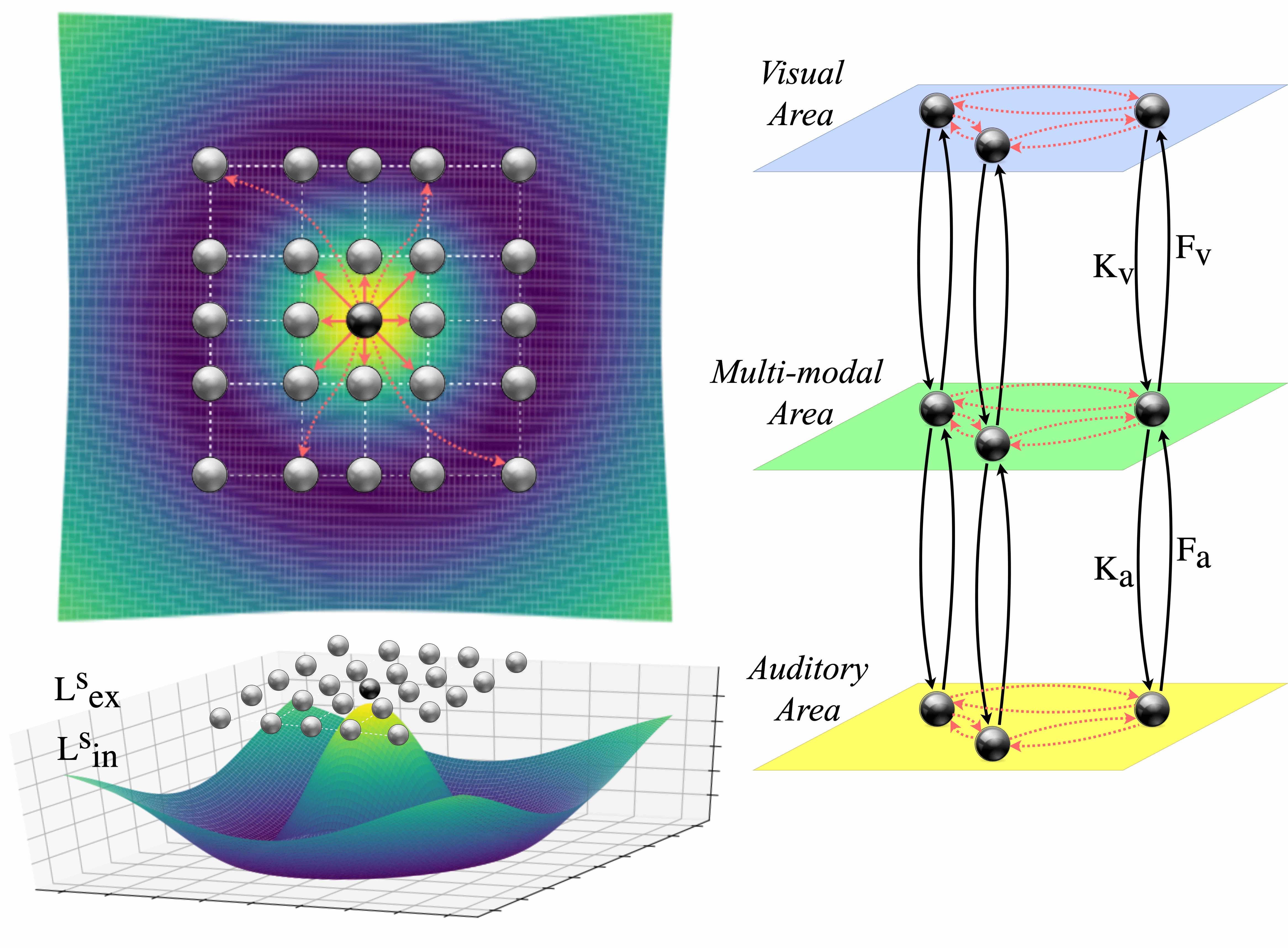}
            \caption{Overall superior colliculus fusion (SCF) layer structure, where the synapses of the central darker neuron (representing neuron $ij$ in a $N \times M$ neural area) are modelled after a Mexican Hat wavelet. Each neuron is laterally linked to its area $s$ neighbors by excitatory $ex$ and inhibitory $in$ synapses (red arrows). The described spatially conditional connection between uni-modal neurons (visual/auditory) and their multi-modal counterparts is shown on the right.}
            \label{fig:connections}
        \end{figure}

        The proposed SC fusion layer, henceforth abbreviated to SCF, is composed by two upstream uni-modal neural areas corresponding to the audio and visual modalities, plus the additional downstream multi-modal area in charge of audiovisual integration. Each upstream area comprises $N \times M$ basic neuron units interlinked within that same area, but not between upstream areas, and topologically organized to communicate inputs to corresponding downstream neurons whose receptive fields (RF)---kernels---are in the same spatial positions. The downstream area in turn, also made up of $N \times M$ neurons, feeds back to each upstream area in the same neuron spatial fashion. This serves to emulate how proximal biological neurons respond to stimuli in proximal position of space, and allows for visual area neurons to indirectly excite audio area neurons or vice-versa through the multi-modal area. Lateral synapses between intra-area neurons are modelled after a mexican hat (Ricker) wavelet, so they may be both excitatory (central region) and inhibitory (surrounding region). The overall described structure is depicted in Figure~\ref{fig:connections}.
        
        Considering a neural area $s \in S=\{a, v, m\}$, pertaining to the auditory $a$, visual $v$ and multi-modal $m$ factors. Asserting $s$ to be composed of $N \times M$ neurons as previously described, each neuron $ij$ or $hk$ receives a composite input $u^{s}_{ij}[n]$, which is described by Equation (\ref{eq:input}). Here $i,j,h $ and $k$ denote iterators over $N \times M$ as such: $i, h=1, \ldots ,N; j,k=1, \ldots ,M$.
        
        \begin{equation}
            \hspace{-0.19cm}
            u^{s}_{ij}[n]= \left\{
            \begin{array}{ll}
                  r^{s}_{ij}[n] + l^{s}_{ij}[n] + f^{s}_{ij}[n], \hspace{0.15cm} s = \{a,v\} \vspace{0.5cm}\\ 
                  k^a \cdot z^{a}_{ij}[n] + k^v \cdot z^{v}_{ij}[n] + l^{s}_{ij}[n], \hspace{0.15cm} s = m\\
            \end{array} 
            \right.  
            \label{eq:input}
        \end{equation}
where $r^{s}_{ij}[n]$ denotes the input at neuron $ij$ triggered by an external stimulus, $l^{s}_{ij}[n]$ represents the input caused by lateral synapses between intra-area neurons, $f^{s}_{ij}[n]$ is the feedback input from the multi-modal SC neurons up to the uni-modal neurons, and $z^{s}_{ij}[n]$ describes an arbitrary neuron's activity whose continuous differential model is discretized in our implementation as the signal in Equation (\ref{eq:activity}) for any $s \in S$.
        
        \begin{equation}
            \hspace{-0.18cm}
            \begin{array}{ll}
                \tau_s z^{s}_{ij}[n + 1] = (\tau_s - 1)z^{s}_{ij}[n] + \phi(p^s(u^{s}_{ij}[n] - \vartheta^s))
            \end{array} 
            \label{eq:activity}
        \end{equation}

        For such a response $\tau_s$ refers to the speed with which a neuron responds to some stimulus, and $\phi$ is the sigmoidal activation of some input $u^{s}_{ij}[n]$ normalized by $\vartheta^s$ and $p^s$---respectively the input value at the central point of neuron activity and its slope. As a consequence, the input to a multi-modal neuron can be seen as the weighted sum of the input activities of each of its spatial uni-sensory counterparts ($k^a$ and $k^v$ being inter-area synaptic connection intensity) plus the area's own lateral synapses.
        
        The laterally synaptic input received by some neuron $ij$ of an area $s$ is defined simply as the weighted sum of the neighboring neurons activities, following (\ref{eq:lateral}). Here $L^s_{ij,hk}$ represents the strength of the synaptic connection from neuron $hk$ to neuron $ij$, which follows the mexican hat wavelet in (\ref{eq:mhat}) as previously explained.

        \begin{equation}
            \begin{array}{ll}
                l^{s}_{ij}[n] = \sum_{h,k}L^s_{ij,hk} \cdot z^{s}_{hk}[n],   &   s \in S
            \end{array} 
            \label{eq:lateral}
        \end{equation}
        
        \begin{equation}
            \hspace{-0.3cm}
            \begin{array}{ll}
                L^s_{ij,hk} = L_{ex}^s \cdot e^{-\frac{[d_x^2 + d_y^2]}{2 \cdot (\sigma_{ex}^s)^2}} - L_{in}^s \cdot e^{-\frac{[d_x^2 + d_y^2]}{2 \cdot (\sigma_{in}^s)^2}},  &   s \in S
            \end{array} 
            \label{eq:mhat}
        \end{equation}
where the $L_{ex}, \sigma_{ex}$ and $L_{in}, \sigma_{in}$ parameters define the strength and spatial disposition of excitatory and inhibitory lateral synapses, respectively. As for $d_x$ and $d_y$ these refer to the inter-area distances between vertically and horizontally intervening neurons, being calculated circularly to prevent border issues.
        
        In terms of the external stimulus input at each neuron $r^{s}_{ij}[n]$, this was obtained through the inner product of the stimulus $I_s$ with the neuron's own receptive field $R_{ij}^s$, made trainable and adaptive at each epoch, as in (\ref{eq:inner}). This was also the case for the feedback input $f^{s}_{ij}[n]$, being calculated as the inner product of a multi-modal neuron's activity $z^{m}_{ij}[n]$ and $F_s$, the strength of the feedback synaptic connection, as in (\ref{eq:inner2}).
        
        \begin{equation}
            \begin{array}{ll}
                r^{s}_{ij}[n] = R_{ij}^s \cdot I_s,  &   s = \{a,v\}
            \end{array} 
            \label{eq:inner}
        \end{equation}
        
        \begin{equation}
            \begin{array}{ll}
                f^{s}_{ij}[n] = F_s \cdot z^{m}_{ij}[n],  &   s = \{a,v\}
            \end{array} 
            \label{eq:inner2}
        \end{equation}

        The motivation behind the proposed layer structure comes from the need to effectively fuse single modality embeddings. The ideal result is an enhanced multi-modal embedding of the most useful audio-video correlations and features. This is achieved thanks to the SC emulation area which upon receiving a multi-modal stimulus reinforces each of its uni-sensory fractions upstream to spatially correlated neuron regions. This concept has greater reach than the one demonstrated on this paper, as there is absolutely no restrictions in terms of number of modalities for fusion---a posture analysis neural area could be added by extending (\ref{eq:input}), behaving similarly to the auditory and visual areas. Moreover, prior sub-fusion areas could also be integrated in the model to combine more closely correlated information---the vision neural area could be divided in two in the case of using a stereo rather than mono visual capture technique. Finally, the structure's described non-trainable parameters were made fine tunable to application specific goals but we kept their default values in our implementation, being supported by the literature pointed in \cite{Ursino}.

    \subsection{Audio Clips and VGGVox}
        
        In \cite{nagrani}, the {VGGVox} model was first introduced as a modified VGG-M CNN\linebreak \mbox{architecture \cite{Chatfield}} aimed at speaker recognition through examination of spectral representations of audio clips. Given its extensive training with over 100,000 utterances by 7000+ distinct speakers, the observed excellent performance has been recognized as a state-of-the-art standard in recent research. Thus, the model is ideal for extrapolating speaker specific cues and discern multiple distinct speakers in an active speaker detection scenario.
        
        In terms of actual audio clips to be analyzed, they should be required to undergo minimal preprocessing only while still being able to progress through a machine learning model. In the past, many techniques have been proposed to obtain audio representations which achieve this goal such as the known Mel-frequency cepstral coefficients \cite{Mermelstein}. Despite their success, such techniques are noted to introduce increased overhead and require a manual tuning which somewhat invalidates the posterior model's positive performance. Therefore for each raw audio clip in our approach, a sliding Hamming window of width 25ms was applied with a step of 10ms in order to generate a respective narrowband spectrogram, as well as means and variances being normalized at each frequency bin of the spectrum. These operations were the single ones performed on the audio data, following {VGGVox}'s preprocessing stage.

    \subsection{Image Sequences and RN-LSTM}

        The {ResNet-50} residual network architecture is one based on the deep residual learning process proposed in the work \cite{He}, which may shortcut the connections between the inputs and outputs of a network's intermediate layers---creating identity mappings. This procedure is performed in order to attempt approximation of residual functions in lieu of fitting a more complex mapping, thanks to the ease of learning advantage caused by occasional skipping of layers considered less relevant. Through skipping, increasingly deeper models may be created without the inherent accuracy degradation previously observed in \cite{He_2} and still benefiting from improved generalization. In addition, the identity mappings also imply a loss similar to that of their shallow non-residual equivalent models. For these reasons, residual networks should be capable of achieving the level of generalization necessary for extracting activity/inactivity cues from face images. 
        
        Whilst residual networks may be suitable for the task, these were not originally trained for facial analysis. Furthermore, ASD approaches must unavoidably deal with human faces in real world interactions which are obviously unconstrained by pose or expression, besides potentially encompassing several and extremely diversified physical attributes. As a means of dealing with these face issues and other common image problems (e.g., varying light/exposure) while still benefiting from the advantages of residual architectures, the {ResNet-50} model trained with {VGGFace2} made available by \cite{Cao} was chosen for our~approach.

        Considering how little to no conclusions pertaining to speech activity may be drawn from solo face images, analyzing the sequential patterns between them becomes a requirement for success. Undoubtedly, recurrent neural networks and more specifically LSTMs are highly suitable for approaching that requirement given their demonstrated past success when dealing with the temporal correlation of facial embeddings \cite{Kankanamge,Xu}. Thus a wide double-layered LSTM was appended to the trained {ResNet-50} architecture creating the mentioned RN-LSTM. This new recurrent section was trained freely by freezing the weights of its preceding {ResNet-50} layers.
        
\begin{figure}[H]
            \includegraphics[width=\columnwidth]{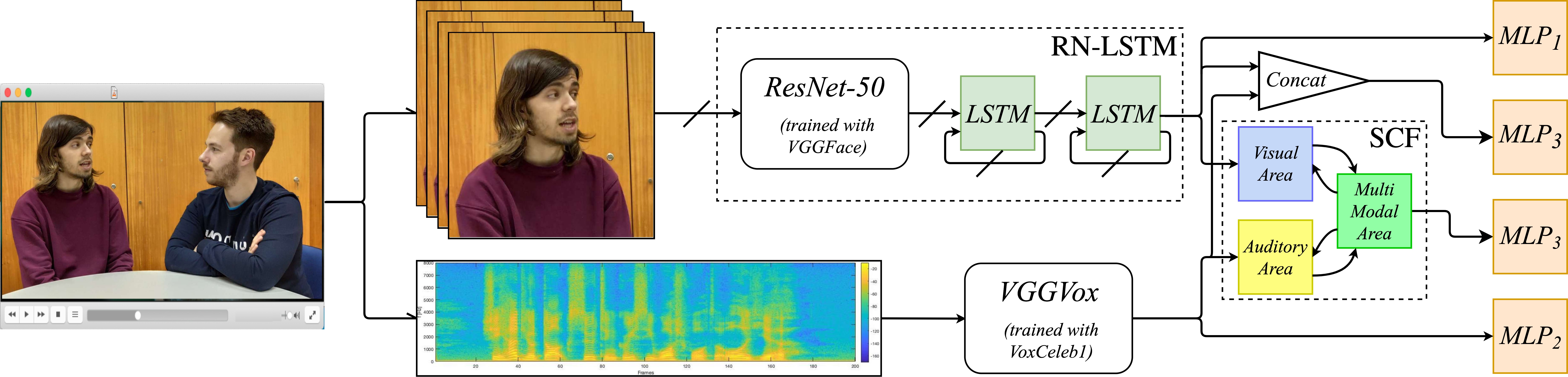}
            \caption{Structure overview of the proposed model. From a video, the image sequence of a face is extracted as well as the corresponding audio. A narrowband spectrogram is generated from the audio and progressed through {VGGVox}. The image sequence is progressed through the RN-LSTM 
            without any preprocessing. The obtained audio and visual embeddings are then classified either separately or merged by concatenation or the SCF layer.}
            \label{fig:diagram}
        \end{figure}

        Evidently the embeddings used for the concatenation approach as well as the input auditory + visual stimulus for the novel SCF layer method were obtained using this RN-LSTM module and the previously described {VGGVox} network. A comprehensive diagram of the overall structure is shown in Figure~\ref{fig:diagram}.
        
\section{Experiments}

    The experimental portion of our work consisted of two distinct parts: (1) Data preparation, given how the condition of the available data was not suitable for proper immediate testing; (2) modality experimentation, where each modality was tested first individually and then fused together. The ultimate goal of each experiment was to recognize, out of $N$ conversation participants, which $M\leq N$ were actively speaking at any moment.

    \subsection{Data Preparation}

        The evaluation of the proposed ASD system required a considerably large amount of data, preferably with a high degree of heterogeneity and with respect to natural conversations. Yet as previously noted, no single dataset encompasses all these characteristics and so the testing phase examined data from two sources: The {Columbia} dataset \cite{Chakravarty_2}; and the 20 videos considered acceptable from the {AVA-ActiveSpeaker} dataset \cite{Roth}. All of the latter's sequences were already under the 10 second mark and so were left untouched. The former's sequences however were originally 136 and ranged from less than 1 second to several minutes in terms of duration. Consequently and considering the standard 30 fps frame rate of the original video, all instances above the 30 frame mark were segmented into 30 frame long sequences. This was performed in order to obtain a much larger number of instances and take better advantage of all the available data for training and testing. Sequences from both datasets below the 30 frame mark were discarded, as instances shorter than 1 second were considered to be similar to static images in terms of unusefulness for active speaker detection. It is also important to note that each segment during training and testing contained a single label (e.g., ``SPEAKING''), yet the condition may very well vary throughout (e.g., a ``SPEAKING'' segment may contain small portions of silence or vice-versa) as speech activity is heterogeneous in nature and somewhat hard to classify consistently. Nonetheless, the models were always given approximately the same number of instances for each class, so that distribution was balanced.
        
        Given how the parsed sequences of {AVA-ActiveSpeaker} were of varying lengths, masking was also employed in order to enable all available sequences to be accepted by the RN-LSTM. Considering this, training was independent of frame window size, making it arbitrary (within 1 and 10 seconds) for model applications. As for the {AVA-ActiveSpeaker} sequences pertaining to the third label ``SPEAKING\_NOT\_AUDIBLE'', these were considered for the visual only portion of the experiments and disregarded for the rest. The overall characteristics of the evaluated data are specified in Table~\ref{tab:data_chars}.

        \begin{specialtable}[H] 
            \caption{Overview of the data used in the experiments. The {Unique Facial Trackings} factor is provided in place of number of speakers considering that {AVA-ActiveSpeaker} does not inform on speaker amount. Same person has been manually verified to not exist in two different videos.}
            \label{tab:data_chars}
            \setlength{\cellWidtha}{\columnwidth/5-2\tabcolsep+0.0in}
\setlength{\cellWidthb}{\columnwidth/5-2\tabcolsep+0.0in}
\setlength{\cellWidthc}{\columnwidth/5-2\tabcolsep+0.0in}
\setlength{\cellWidthd}{\columnwidth/5-2\tabcolsep+0.0in}
\setlength{\cellWidthe}{\columnwidth/5-2\tabcolsep+0.0in}
\scalebox{1}[1]{\begin{tabularx}{\columnwidth}{>{\PreserveBackslash\centering}m{\cellWidtha}>{\PreserveBackslash\centering}m{\cellWidthb}>{\PreserveBackslash\centering}m{\cellWidthc}>{\PreserveBackslash\centering}m{\cellWidthd}>{\PreserveBackslash\centering}m{\cellWidthe}}
\toprule
                \textbf{Dataset} & \textbf{Sequences} & \textbf{Frames} & \textbf {Unique  Facial Trackings}  & \textbf{Image  Conditions} \\ \midrule
                {Columbia} & 4914 & 30 & 34 (6 speakers) & Identical \\
                {AVA-ActiveSpeaker} & 5510 (3 labels)\linebreak 5470 (2 labels) & {[}30, 300{]} & 4321 (20 videos) & Varied \\ \bottomrule
            \end{tabularx}}
        \end{specialtable}

    \subsection{Modality Experimentation}

        Before attempting classification of the fused multi-modal embeddings, an initial testing of each individual modality was performed to assess the extrapolation quality and obtain a term of comparison for later on. Accordingly, multilayer perceptrons (MLP) were attached to the end of the feature extraction layers of {VGGVox} and the developed RN-LSTM, for classification. Additionally, level of model generalization was also assessed by comparing dependence and independence scenarios in terms of speaker for the {Columbia} dataset and in terms of video for {AVA-ActiveSpeaker}---a solution to the unknown speaker amount issue. Essentially, for dependent scenario experiments (closed set) the test data includes instances from speakers who are also represented in the training data, albeit by completely distinct instances. On the contrary, for independent scenario experiments (open set) it is guaranteed that the speakers from the test data are completely different and never before seen by the model during training. These two types of testing allow for a better evaluation of the model's generalization ability, by examining how much the model's performance decreases from dependency to independency. The obtained results are presented in Tables~\ref{tab:columbia_audio_video} and \ref{tab:ava_audio_video}, respectively.

        \begin{specialtable}[H]
            \caption{Audio/video results using the Columbia dataset with respect to {Mean Accuracy\linebreak (Standard~Deviation)}.}
            \label{tab:columbia_audio_video}
           \setlength{\cellWidtha}{\columnwidth/3-2\tabcolsep+0.0in}
\setlength{\cellWidthb}{\columnwidth/3-2\tabcolsep+0.0in}
\setlength{\cellWidthc}{\columnwidth/3-2\tabcolsep+0.0in}
\scalebox{1}[1]{\begin{tabularx}{\columnwidth}{>{\PreserveBackslash\centering}m{\cellWidtha}>{\PreserveBackslash\centering}m{\cellWidthb}>{\PreserveBackslash\centering}m{\cellWidthc}}
\toprule
             & \textbf{Speaker Dependent} & \textbf{Speaker Independent} \\ \midrule
           $VGGVox + MLP_2$\linebreak (Audio Only) & 47.37 (0.52) [\%] & 17.34 (9.14) [\%] \\
           $RN$-$LSTM + MLP_1$\linebreak (Video Only)& 97.68 (0.39) [\%] & 64.89 (14.44) [\%] \\ \bottomrule
            \end{tabularx}}
        \end{specialtable}
        \vspace{-6pt}
        \begin{specialtable}[H]
            \caption{Audio/video results using the AVA-ActiveSpeaker videos with respect to {Mean Accuracy (Standard Deviation)}.}
            \label{tab:ava_audio_video}
            \setlength{\cellWidtha}{\columnwidth/3-2\tabcolsep+0.0in}
\setlength{\cellWidthb}{\columnwidth/3-2\tabcolsep+0.0in}
\setlength{\cellWidthc}{\columnwidth/3-2\tabcolsep+0.0in}
\scalebox{1}[1]{\begin{tabularx}{\columnwidth}{>{\PreserveBackslash\centering}m{\cellWidtha}>{\PreserveBackslash\centering}m{\cellWidthb}>{\PreserveBackslash\centering}m{\cellWidthc}}
\toprule
             & \textbf{Video Dependent} & \textbf{Video Independent} \\ \midrule
            $VGGVox + MLP_2$\linebreak (Audio Only) & 87.73 (0.87) [\%] & 84.96 (1.78) [\%] \\
            $RN$-$LSTM + MLP_1$\linebreak (Video Only) & 78.68 (1.79) [\%] & 72.48 (2.17) [\%] \\ \bottomrule
           \end{tabularx}}
        \end{specialtable}

        The novel SCF layer, as well as the entirety of the implemented experiments, were realized using the Keras v2.4.3 framework \cite{Keras} with Tensorflow backend. Test runs were performed on a Nvidia RTX 2080 GPU with 8 GB memory. In order to assess its superiority against state-of-the-art standards and suitability to the task at hand, we compared its performance to a naive baseline. This was composed of an audiovisual concatenated embedding array fed directly to a multi-layer perceptron, as seen in Figure~\ref{fig:diagram}. The results of this experiment and those obtained with the SCF layer alternative are shown in \mbox{Tables~\ref{tab:columbia_audiovisual} and \ref{tab:ava_audiovisual}} for the Columbia and AVA datasets, respectively.
        
        \begin{specialtable}[H]
            \caption{Audiovisual results using the Columbia dataset with respect to {Mean Accuracy\linebreak (Standard Deviation)}.}
%
            \label{tab:columbia_audiovisual}
            \setlength{\cellWidtha}{\columnwidth/3-2\tabcolsep+0.0in}
\setlength{\cellWidthb}{\columnwidth/3-2\tabcolsep+0.0in}
\setlength{\cellWidthc}{\columnwidth/3-2\tabcolsep+0.0in}
\scalebox{1}[1]{\begin{tabularx}{\columnwidth}{>{\PreserveBackslash\centering}m{\cellWidtha}>{\PreserveBackslash\centering}m{\cellWidthb}>{\PreserveBackslash\centering}m{\cellWidthc}}
\toprule
             & \textbf{Speaker Dependent} & \textbf{Speaker Independent} \\ \midrule
           $Concat + MLP_3$\linebreak (Baseline) & 97.60 (0.49) [\%] & 41.78 (15.46) [\%] \\
            $SCF + MLP_3$\linebreak (New Approach) & 98.53 (0.21) [\%] & 58.50 (7.71) [\%] \\ \bottomrule
 \end{tabularx}}
        \end{specialtable}
        \vspace{-6pt}
        
        \begin{specialtable}[H]
            \caption{Audiovisual results using the AVA-ActiveSpeaker videos with respect to {Mean Accuracy (Standard Deviation)}.}
            \label{tab:ava_audiovisual}
             \setlength{\cellWidtha}{\columnwidth/3-2\tabcolsep+0.0in}
\setlength{\cellWidthb}{\columnwidth/3-2\tabcolsep+0.0in}
\setlength{\cellWidthc}{\columnwidth/3-2\tabcolsep+0.0in}
\scalebox{1}[1]{\begin{tabularx}{\columnwidth}{>{\PreserveBackslash\centering}m{\cellWidtha}>{\PreserveBackslash\centering}m{\cellWidthb}>{\PreserveBackslash\centering}m{\cellWidthc}}
\toprule
             & \textbf{Video Dependent} & \textbf{Video Independent} \\ \midrule
           $Concat + MLP_3$\linebreak (Baseline) & 89.10 (0.41) [\%] & 87.52 (2.49) [\%] \\
            $SCF + MLP_3$\linebreak (New Approach) & 91.68 (0.51) [\%] & 88.46 (2.06) [\%] \\ \bottomrule
           \end{tabularx}}
        \end{specialtable}
        
        The SCF layer's neural areas were kept at a $17 \times 17$ dimension, for the sake of computational simplicity. Each of the presented values in all tables was obtained using 5-fold cross validation for increased robustness. Furthermore, the adaptive moment estimation (Adam) optimizer \cite{Kingma} was employed during the training phase of the models for weight adjustment. Batch normalization \cite{Ioffe} was also integrated in the training, in addition to minor dropout \cite{Srivastava}. The number of epochs was maintained constant across transversal testing, having each ideal value been obtained through early stopping mechanism for the three experimental phases (video-only, audio-only and audiovisual).

\section{Discussion}

    As a first observation, it can be noted how the audio-only approach to active speaker detection (Tables~\ref{tab:columbia_audio_video} and \ref{tab:ava_audio_video}) performed surprisingly well for the AVA-ActiveSpeaker videos but was rather unsuccessful for the Columbia dataset. This is attributed to the mentioned data homogeneity of the latter, considering it encompasses a single speaker panel video. The fact that there is constant speech in the video (one speaker picks up after the other) prevents the model from learning even the most basic factor of discerning silence from speech. Hence results matching random decision for speaker dependency but failing with the independent case.
    
    Still regarding the Columbia dataset's homogeneity, it proves to be advantageous for the dependent section of the video-only testing phase, but again not so much for the independent one. Thus the model naturally performs better with previously seen subjects but fails to extrapolate suitable ASD features by training with such a reduced number of speakers. Not unlike the audio testing however, the AVA videos led to a considerably high performance using images only, proving data heterogeneity and speaker abundance to be ultimately better for training as expected.

\begin{figure}[H]
        \includegraphics[width=\columnwidth]{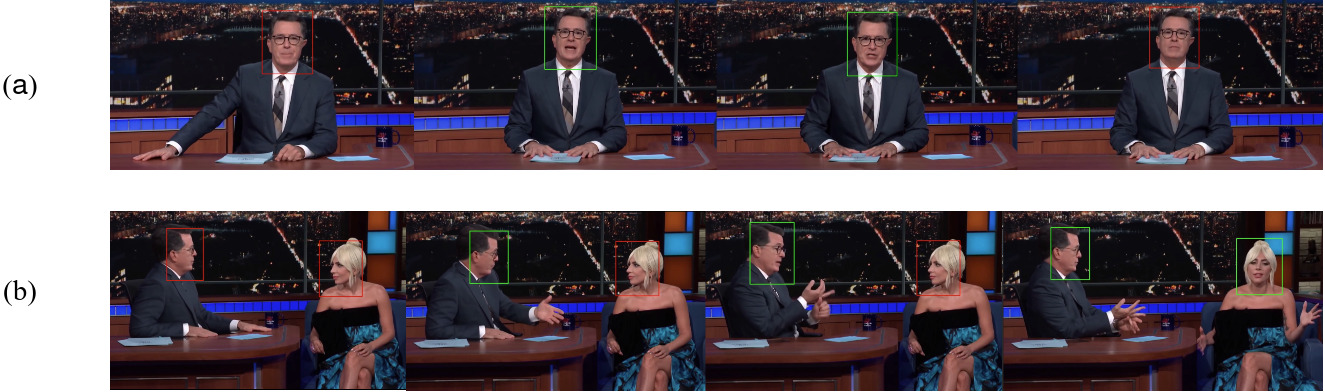}
        \caption{Performance of the trained full model on unseen data: (\textbf{a}) Single speaker scenario. (\textbf{b}) Multiple speaker scenario.}
        \label{fig:f2}
    \end{figure}

    Audiovisual results greatly surpassed expectations, with the baseline concatenation approach already demonstrating state-of-the-art performance for either dataset considered. Increase in terms of standard deviation from the dependent to the independent testing was observed as expected, considering how the model suffers a natural decrease of confidence in its predictions when analyzing previously unseen speakers. Relatively poor results are of course still obtained for the independent part of the Columbia dataset testing due to the constant speech issue, which as expected severely hinders the model's performance. Nonetheless, results are remarkably positive for either concatenation or SCF layer approaches with respect to speaker dependent testing. Moreover, the SCF layer still showed its superiority by majorly improving over concatenation and even surpassing a random choice performance despite having to deal with crippling audio data. As for the AVA video testing, both the concatenation and SCF results beat those presented in the dataset paper \cite{Roth}, because even though testing was carried out with only 20 of the original videos, the performance gain was still exemplary in terms of result excellence. Even more so, the SCF layer successfully improved the already excellent baseline performance. This serves to validate its bio-inspired concept---mutual neuron excitation of spatially proximal multi-modal neural regions for integration of uni-sensory information---here emulated for modality fusion. Undoubtedly even greater performance gains could be achieved if all the layer's hyperparameters were finely tuned to our application, rather than using default values. Nonetheless, results were highly successful and warrant further research in this and other areas encompassing several types of modalities. Finally, the full model performed quite well on a snippet of novel data not part of either datasets used during training and testing. Stills from this snippet are shown in Figure~\ref{fig:f2}. It should be noted that, even though only two speakers are shown in this example, the technique is scalable to an indefinite number of speakers.

    Evidently the technique is not without limitations. In terms of its application to ASD, data with significantly large groups of people can result in degraded efficiency caused by extra image/audio embedding evaluation and fusion happening concurrently. This problem can be mitigated provided the underlying hardware is able to handle the increased computational load. Specifically with the hardware used for our implementation, we found real-time ASD to be possible with negligible delay. Other application limitations are relatively uncommon but can occur for instance when audience noise voices over a silent but moving speaker, making the latter appear active, or when two spatially close speakers move and sound similar. This can be addressed by added preprocessing to remove background noise and non-speech sound as well as further training with more heterogeneous data. Data which should become more readily available as research into the novel topic of ASD advances. As for limitations of the proposed fusion module, it is not a straightforward issue to find ideal values for the coefficients of neuron input and activity. Whilst here values were empirically selected, adapting these into trainable parameters of a neural network would certainly yield better performance.

\section{Conclusions}

    This work proposed a novel modality fusion layer (SCF), bio-inspired by the brain's superior colliculus region and incorporated into a new method for ASD in real-world scenarios. This proposed method deals with audio and video information jointly and makes no assumptions regarding input data. The audiovisual technique was extensively tested against data from two tried and tested ASD databases, specifically the \mbox{Columbia \cite{Chakravarty_2}} and the AVA-ActiveSpeaker \cite{Roth} datasets. The obtained results were compared to those of audio-only and video-only approaches, as well as those of a naive concatenation multi-modal baseline. In any case the SCF technique always demonstrated state-of-the-art performance, surpassing its baseline and uni-modal counterparts. It was shown how mutual neuron excitation of a spatially proximal multi-modal region by uni-modal neurons can be a remarkably successful modality fusion technique.
    
    This observed SCF layer's success in terms of uni-sensory data integration, even despite doing so with default settings, evidently warrants a greater exploration of its capabilities. As such, we further intend to assess whether the increased performance of the SCF layer w.r.t. simple concatenation is even more accentuated when one of the modalities is missing. In addition, we will examine the layer's behavior in terms of not only audiovisual fusion but also combination of data of other natures (e.g., tactile).

\end{paracol}

 \begin{paracol}{2}
 \switchcolumn




\vspace{6pt} 



\authorcontributions{Conceptualization, G.A., N.G. and P.M.; methodology, G.A., N.G. and P.M.; software, G.A.; validation, G.A.; data curation, G.A.; writing---original draft preparation, G.A.; writing---review and editing, N.G. and P.Menezes; supervision, P.M.; All authors have read and agreed to the published version of the manuscript.}

\funding{This research was funded in part by scholarhip 2020.05620.BD of the Fundação para a Ciência e a Tecnologia (FCT) of Portugal, and OE - National funds of FCT/MCTES under project number UIDB/00048/2020.}

\institutionalreview{Not applicable.}

\informedconsent{Not applicable.}

\dataavailability{Publicly available datasets were analyzed in this study. The Columbia dataset video can be found in \url{https://youtu.be/6GzxbrO0DHM}. Annotations are available at \url{http://www.jaychakravarty.com/wp-content/uploads/2016/02/columbiaDatasetGroundTruth.zip}. The full AVA-ActiveSpeaker dataset can be found in \url{https://research.google.com/ava/download.html\#ava_active_speaker_download}. The list of the 20 AVA-ActiveSpeaker videos analyzed in this study is avaiable at \url{https://github.com/gustavomiguelsa/SCF}. No new data were created in this study.} 

\conflictsofinterest{The authors declare no conflict of interest. The funders had no role in the design of the study; in the collection, analyses, or interpretation of data; in the writing of the manuscript, or in the decision to publish the results.}


\abbreviations{The following abbreviations are used in this manuscript:\\

\noindent 
\begin{tabular}{@{}ll}
ASD & Active speaker detection\\
VAD & Voice activity detection\\
TDOA & Time difference of arrival\\
CNN & Convolutional neural network\\
LSTM & Long short term memory\\
RN-LSTM & Resnet-LSTM\\
SC & Superior colliculus\\
SCF & Superior colliculus fusion
\end{tabular}}

\end{paracol}
\reftitle{References}

%



\begin{thebibliography}{999}
\bibitem{nagrani} Nagrani, A.; Chung, J.S.; Zisserman, A. Voxceleb: A large-scale speaker identification dataset. \emph{arXiv} \textbf{2017}, arXiv:1706.08612. 

\bibitem{Ephrat} Ephrat, A.; Mosseri, I.; Lang, O.; Dekel, T.; Wilson, K.; Hassidim, A.; Freeman, W.T.; Rubinstein, M.  Looking to listen at the cocktail party: A speaker-independent audio-visual model for speech separation. ACT Trans. Graph. \textbf{2018}, \emph{37}, 1--11. 

\bibitem{Vajaria} Vajaria, H.; Sarkar, S.; Kasturi, R. Exploring co-occurrence between speech and body movement for audio-guided video localization.\emph{ IEEE Trans.  Circuits  Syst.  Video  Technol}. \textbf{2008},  \emph{18}, 1608–1617.

\bibitem{Tranter} Tranter, S.E.; Reynolds, D.A. An overview of automatic speaker diarization systems. \emph{IEEE Trans. Audio Speech Lang. Process.} \textbf{2006}, \emph{14}, 1557--1565.


\bibitem{Bouquin} le Bouquin-Jeannès, R.; Faucon, G. Study of a voice activity detector and its influence on a noise reduction system. \emph{Speech Comm}. \textbf{1995}, \emph{16}, 245--254.

\bibitem{Liu} Liu, D.; Kubala, F. Online speaker clustering.  In Proceedings of the 2003 IEEE International Conference on Acoustics, Speech, and Signal Processing, 2003. Proceedings. (ICASSP '03), Hong Kong, China,  6--10 April 2003.


\bibitem{Maraboina} Maraboina, S.; Kolossa, D.; Bora, P.K.; Orglmeister, R. Multi-speaker voice activity detection using ICA and beampattern analysis.  In Proceedings of the 2006 14th European Signal Processing Conference, Florence, Italy, 4--8 September 2006; pp. 1--5. 
    
\bibitem{Bertrand} Bertrand, A.; Moonen, M. Energy-based multi-speaker voice activity detection with an ad hoc microphone array. In Proceedings of the 2010 IEEE International Conference on Acoustics, Speech and Signal Processing, Dallas, TX, USA,  14--19 March 2010; pp.~85--88. 

\bibitem{Siatras} Siatras, S.; Nikolaidis, N.; Krinidis, M.; Pitas, I. Visual Lip Activity Detection and Speaker Detection Using Mouth Region Intensities.  \emph{IEEE Trans. Circuits Syst. Video Technol.} \textbf{2009}, \emph{19}, 133--137.

\bibitem{Ahmad}Ahmad, R.; Raza, S.P.; Malik, H. Visual Speech Detection Using an Unsupervised Learning Framework. In Proceedings of the 2013 12th International Conference on Machine Learning and Applications, Miami, FL, USA, 4--7 December 2013; pp. 525--528. 

\bibitem{Stefanov} Stefanov, K.; Sugimoto, A.; Beskow, J.  Look  who’s talking:  Visual identification of the active speaker in multi-party human--robot interaction.  InProc. In Proceedings of the Workshop Advancements in Social Signal Processing for Multimodal Interaction, Tokyo, Japan, 16 November 2016; pp. 22–27. 

\bibitem{Minotto} Minotto, V.P.; Jung, C.R.; Lee, B. Simultaneous-Speaker Voice Activity Detection and Localization Using Mid-Fusion of SVM and HMMs.  \emph{IEEE Trans. Multimed.} \textbf{2014}, \emph{16}, 1032--1044.

\bibitem{Cutler} Cutler, R.; Davis, L. Look who's talking: speaker detection using video and audio correlation. In Proceedings of the 2000 IEEE International Conference on Multimedia and Expo. ICME2000. Proceedings. Latest Advances in the Fast Changing World of Multimedia (Cat. No.00TH8532), New York, NY, 30 July--2 August 2000;  Volume 3, pp. 1589--1592. 

\bibitem{Chakravarty} Chakravarty, P.; Mirzaei, S.; Tuytelaars, T.; Vanhamme, H. Who’s speaking? audio-supervised classification of active speakers in video. In Proceedings of the 2015 ACM on International Conference on Multimodal Interaction, Seattle, WA, USA, 9~November~2015. 

\bibitem{Chakravarty_2} Chakravarty, P.; Tuytelaars, T.  Cross-modal supervision for learning active speaker detection in video.  In \emph{European Conference on Computer Vision}; Springer: Cham, Switzerland, 2016; pp. 1–5.

\bibitem{Stefanov_2} Stefanov, K.; Beskow, J.; Salvi, G. Vision-based active speaker detection in multiparty interaction. In Proceedings of the GLU 2017 Inter-national Workshop on Grounding Language Understanding, Stockholm, Sweden, 25 August 2017; pp.47–51. 

\bibitem{Fukushima}  Fukushima. K. Neocognitron: A self-organizing neural network model for a mechanism of pattern recognition unaffected by shift in position. \emph{Biol. Cybern.} \textbf{1980},  \emph{36}, 93--202.

\bibitem{Hubel} Hubel, D.H.; Wiesel, T.N. Receptive fields and functional architecture of monkey striate cortex. \emph{J. Physiol.} \textbf{1968}, \emph{195} , 215--243.

\bibitem{Stefanov_3} Stefanov, K.; Beskow, J.; Salvi, G. Self-Supervised Vision-Based Detection of the Active Speaker as Support for Socially-Aware Language Acquisition. \emph{IEEE Trans. Cogn. Dev. Syst.} \textbf{2019}, \emph{12}, 250--259.

\bibitem{Ren} Ren, J.; Hu, Y.; Tai, Y.-W.; Wang, C.; Xu, L.; Sun, W.; Yan, Q. Look, listen and learn---A multimodal LSTM for speaker identification. In Proceedings of the Thirtieth AAAI Conference on Artificial Intelligence, Phoenix, AR, USA, 12--17 February 2016; pp. 3581–3587. 

\bibitem{Cech} Cech, J.; Mittal, R.; Deleforge, A.; Sanchez-Riera, J.X. Alameda-Pineda and R. Horaud. Active-speaker detection and localization with microphones and cameras embedded into a robotic head.  In Proceedings of the 2013 13th IEEE-RAS International Conference on Humanoid Robots (Humanoids), Atlanta, GA, 15--17 October 2013; pp. 203-210. 

\bibitem{Gebru} Gebru, I.D.; Alameda-Pineda, X.; Horaud, R.; Forbes, F. Audio-visual speaker localization via weighted clustering. In Proceedings of the 2014 IEEE International Workshop on Machine Learning for Signal Processing (MLSP), Reims, France, 21--24 September 2014; pp. 1-6. 

\bibitem{Hoover} Hoover, K.; Chaudhuri, S.; Pantofaru, C.; Sturdy, I.; Slaney, M. Using audio-visual information to understand speaker activity: Tracking active speakers on and off screen. In Proceedings of the 2018 IEEE International Conference on Acoustics, Speech and Signal Processing (ICASSP), Calgary, AB, Canada, 15--20 April  2018; pp. 6558--6562. 

\bibitem{Jegou} Jegou, H.; Perronnin, F.; Douze, M.; Sanchez, J.; Perez, P.; Schmid, C. Aggregating Local Image Descriptors into Compact Codes; In \emph{IEEE Transactions on Pattern Analysis and Machine Intelligence}, vol. 34, no. 9, pp. 1704-1716, Sept. 2012.


\bibitem{Roth} Roth, J.; Chaudhuri, S.; Klejch, O.; Marvin, R.; Gallagher, A.C.; Kaver, L.; Ramaswamy, S.; Stopczynski, A.; Schmid, C.; Xi, Z.; et al. AVA-ActiveSpeaker: An Audio-Visual Dataset for Active Speaker Detection. \emph{arXiv} \textbf{2019}, arXiv:abs/1901.01342.

\bibitem{He} He, K.; Zhang, X.; Ren, S.; Sun, J. Deep residual learning for image recognition.  In Proceedings of the Conference on Computer Vision and Pattern Recognition,  Las Vegas, NV, USA, 27--30 June 2016; pp. 770–778. 

\bibitem{Cao} Cao, Q.; Shen, L.; Xie, W.; Parkhi, O.; Zisserman, A. VGGFace2: A Dataset for Recognising Faces across Pose and Age. In Proceedings of the IEEE International Conference on Automatic Face \& Gesture Recognition, Xi'an, China, 15--19 May 2018. 

\bibitem{Keras_vggface} The Keras-VGGFace Package.  Available online: \url{https://pypi.org/project/keras-vggface/}  (accessed on: 2020 November 23).

\bibitem{Ursino} Ursino, M.; Cuppini, C.; Magosso, E.; Serino, A.; di Pellegrino, G. Multisensory integration in the superior colliculus: a neural network model. \emph{J. Comput. Neurosci.} \textbf{2008},  \emph{26}, 55–73.

\bibitem{Chatfield} Chatfield, K.; Simonyan, K.; Vedaldi, A.; Zisserman, A. Return of the devil in the details: Delving deep into convolutional nets. In Proceedings of the British Machine Vision Conference, Nottingham, UK, 1--5 September 2014. 

\bibitem{Mermelstein}  Mermelstein, P. Distance measures for speech recognition, psychological and instrumental. In \emph{Pattern Recognition and Artificial Intelligence};  Chen, C.H.;  Ed.; Academic: New York, NY, USA, 1976;  pp. 374–388.

\bibitem{He_2} He, K.; Sun, J. \emph{Convolutional Neural Networks at Constrained Timecost}; CVPR: 2015. 

\bibitem{Kankanamge} Kankanamge, S.; Fookes, C.; Sridharan, S. Facial analysis in the wild with LSTM networks.  In Proceedings of the 2017 IEEE International Conference on Image Processing (ICIP), Beijing, China, 17--20 September 2017; pp. 1052--1056. 

\bibitem{Xu} Xu, Z.; Li, S.; Deng, W. Learning temporal features using LSTM-CNN architecture for face anti-spoofing. In Proceedings of the 2015 3rd IAPR Asian Conference on Pattern Recognition (ACPR), Kuala  Lumpur, Malaysia, 3--6 November 2015; pp. 141--145. 



\bibitem{Keras} Chollet, F. Keras. 2015. Available online:  \url {https://keras.io } (accessed on: 2020 November 23). 

\bibitem{Kingma} Kingma, D.P.; Ba, J. Adam: A method for stochastic optimization. \emph{arXiv} \textbf{2014}, arXiv:1412.6980.

\bibitem{Ioffe} Ioffe, S.; Szegedy, C. Batch normalization: Acceleratingdeep network training by reducing internal covariate shift.  \emph{arXiv} \textbf{2015}, arXiv:1502.03167.

\bibitem{Srivastava} Srivastava, N.; Hinton, G.; Krizhevsky, A.; Sutskever, I.; Salakhutdinov, R. Dropout: A simple way to prevent neural networks from overfitting. \emph{J. Mach. Learn. Res.} \textbf{2014}, \emph{15}, 1929–1958.
\end{thebibliography}
\end{document}